\documentclass{article}

\usepackage{PRIMEarxiv}

\usepackage[utf8]{inputenc} 
\usepackage[T1]{fontenc}    
\usepackage{hyperref}       
\usepackage{url}            
\usepackage{booktabs}       
\usepackage{amsfonts}       
\usepackage{nicefrac}       
\usepackage{microtype}      
\usepackage{lipsum}
\usepackage{fancyhdr}       
\usepackage{graphicx}       
\graphicspath{{media/}}     
\usepackage{amsmath}
\usepackage{algorithm}
\usepackage{algorithmic}

\usepackage{subcaption}    

\title{ViLLa: A Neuro-Symbolic approach for Animal Monitoring
}

\author{
  Harsha Koduri \\
  \\
  \texttt{harsha.koduri123@gmail.com} \\
\
}

\begin{document}
\maketitle

\begin{abstract}
  Monitoring animal populations in natural environments requires systems that can interpret both visual data and human language queries. This work introduces ViLLa (Vision-Language-Logic Approach), a neuro-symbolic framework designed for interpretable animal monitoring. ViLLa integrates three core components: a visual detection module for identifying animals and their spatial locations in images, a language parser for understanding natural language queries, and a symbolic reasoning layer that applies logic-based inference to answer those queries. Given an image and a question such as “How many dogs are in the scene?” or “Where is the buffalo?”, the system grounds visual detections into symbolic facts and uses predefined rules to compute accurate answers related to count, presence, and location. Unlike end-to-end black-box models, ViLLa separates perception, understanding, and reasoning, offering modularity and transparency. The system was evaluated on a range of animal imagery tasks and demonstrates the ability to bridge visual content with structured, human-interpretable queries.

\noindent\textbf{Code available:} \url{https://github.com/HarshaKoduri123/ViLLa}
\end{abstract}


\section{Introduction}

Monitoring animal populations is essential for ecological research, biodiversity conservation, and habitat management. Accurate insights into animal presence, movement, and population density enable researchers to understand ecosystem dynamics and detect early signs of environmental change. Traditional methods, such as manually reviewing camera trap images or aerial footage, are time-consuming and labor-intensive, limiting both the scale and frequency of analysis. As the volume of visual animal data continues to grow, there is an increasing need for systems capable of automated, flexible, and interpretable analysis.

Recent advances in computer vision—particularly convolutional neural networks (CNNs) \cite{NIPS2012_c399862d, he2015deepresiduallearningimage, szegedy2014goingdeeperconvolutions} and object detection models such as YOLO \cite{reis2024realtimeflyingobjectdetection}—have significantly improved the accuracy and speed of animal detection in images. These tools are effective at identifying and localizing animals in field data collected via drones or static cameras. However, they typically produce raw outputs, such as bounding boxes and class labels, without offering structured answers to high-level queries like “How many zebras are there?” or “Is there a rhino in the image?” Addressing such questions often requires complex, custom post-processing, which reduces interpretability and limits the usability of these models in practical ecological decision-making.

To address these challenges, recent work has explored integrating visual recognition with natural language processing \cite{vaswani2023attentionneed}, leading to the development of vision-language or multimodal models \cite{radford2021learningtransferablevisualmodels, 9710099, li2023lvitlanguagemeetsvision}. While these systems can associate textual queries with image content, they often depend on large, opaque architectures that are difficult to interpret, modify, or validate. An alternative approach involves symbolic reasoning, where both image content and user queries are expressed in structured, logic-based representations. This enables more transparent, rule-driven inference and easier integration of domain-specific knowledge.

This paper introduces ViLLa (Vision-Language-Logic Approach), a modular framework that combines visual detection, language parsing, and symbolic reasoning to support interpretable animal monitoring. Unlike monolithic models, ViLLa separates perception, understanding, and reasoning into distinct, explainable components. A visual detection module extracts symbolic facts (e.g., animal type and location), a language parser translates user queries into structured representations, and a logic engine applies rule-based reasoning to produce precise answers. The framework supports a range of tasks, including counting animals, checking for presence, and identifying spatial locations.

By grounding both visual and textual inputs in a shared symbolic space, ViLLa enables interpretable and task-specific reasoning over complex wildlife scenes. This neuro-symbolic design improves transparency, debuggability, and adaptability, providing a practical tool for conservation scientists and field ecologists. Through a series of wildlife case studies, the framework demonstrates its ability to respond to a diverse range of ecological queries, offering a lightweight and modular alternative to traditional black-box approaches.

\section{Related Work}

\subsection{Vision Models}

Traditional computer vision approaches have relied heavily on convolutional neural networks (CNNs) \cite{NIPS2012_c399862d} \cite{he2015deepresiduallearningimage} \cite{szegedy2014goingdeeperconvolutions} for tasks such as classification and object recognition. These architectures extract hierarchical spatial features and have been the backbone of many early visual recognition systems. As object detection gained prominence, models like Faster R-CNN \cite{ren2016fasterrcnnrealtimeobject}, SSD \cite{Liu_2016}, and YOLO \cite{reis2024realtimeflyingobjectdetection} were introduced to localize multiple objects within an image. These detection models significantly improved performance in real-time applications and fine-grained visual categorization. 

More recently, Vision Transformers (ViTs) \cite{dosovitskiy2021imageworth16x16words} have reshaped the field by treating image patches as tokens and applying self-attention mechanisms, allowing for more global context modeling. This has led to competitive results on tasks like image classification and detection. Variants of ViTs have been adapted for dense prediction \cite{ranftl2021visiontransformersdenseprediction}, instance segmentation \cite{zhang2022segvitsemanticsegmentationplain}, and object tracking \cite{Zhang_2021_ICCV}, offering a scalable alternative to CNN-based detectors, especially in scenarios requiring semantic reasoning. 

\subsection{Language Models}

In early natural language processing systems, models such as Recurrent Neural Networks (RNNs) \cite{650093}, Gated Recurrent Units (GRUs)\cite{chung2014empiricalevaluationgatedrecurrent}, and Long Short-Term Memory (LSTM) \cite{10.1162/neco.1997.9.8.1735} networks were used to capture sequential dependencies. These architectures enabled language understanding in tasks such as translation, summarization, and question answering. However, their limited ability to scale and capture long-range dependencies motivated the shift toward transformer-based models. 

The introduction of transformer architectures led to a wave of large language models (LLMs), such as BERT\cite{devlin2019bertpretrainingdeepbidirectional}, GPT\cite{brown2020languagemodelsfewshotlearners}, and LLaMA \cite{touvron2023llamaopenefficientfoundation}, which offer contextualized representations and have shown strong performance across a wide range of language tasks. Fine-tuning techniques allow these models to adapt to domain-specific tasks with relatively small labeled datasets. Other transformer variants like MPNet\cite{song2020mpnetmaskedpermutedpretraining} have introduced masked and permuted pretraining objectives for improved contextual understanding and representation learning.

\subsection{Symbolic AI}

Symbolic Artificial Intelligence (AI) , often referred to as classical AI, is rooted in the use of formal logic systems  \cite{info11030167}, rule-based reasoning \cite{kirrane2024rules}\cite{FRYE1995483}, and structured symbolic representations. It enables machines to perform deductive reasoning by manipulating symbols according to explicitly defined rules. Symbolic AI has traditionally been used in expert systems, automated theorem proving, and knowledge-based decision-making. Its strength lies in its interpretability and the ability to encode domain knowledge with precision. However, symbolic systems often struggle with perception-based tasks and lack flexibility in dealing with ambiguous or noisy data, which limits their standalone application in real-world scenarios that involve unstructured information.

\subsection{Neuro-Symbolic AI}

Neuro-symbolic \cite{10.3233/AIC-210084} methods aim to merge the strengths of data-driven perception with structured reasoning. While conventional machine learning techniques are proficient at recognizing patterns in images and understanding basic language input, they often struggle when interpretability or logic-based inference is required. Symbolic systems, in contrast, rely on rule-based logic to produce deterministic outputs and can represent knowledge in a transparent and verifiable way.

In practical terms, neuro-symbolic systems extract structured information—such as object classes, counts, or spatial coordinates—from raw data and then use formal rules to make decisions or answer queries. This hybrid approach allows systems to process complex inputs like photographs and text, and return answers that can be logically traced. Instead of relying entirely on large-scale training, these systems can work effectively with domain-specific rules and moderate data.

In the context of this work, this approach is applied to animal monitoring by combining object detections from images with text-derived queries, converting them into symbolic representations. Logic-based reasoning is then used to determine answers such as the number of animals, whether a particular species is present, or its location. This method offers a structured, interpretable framework suited for ecological monitoring and conservation use cases.

\section{Methodology}

\begin{figure}[!t]
  \begin{center}
    \includegraphics[width=\textwidth]{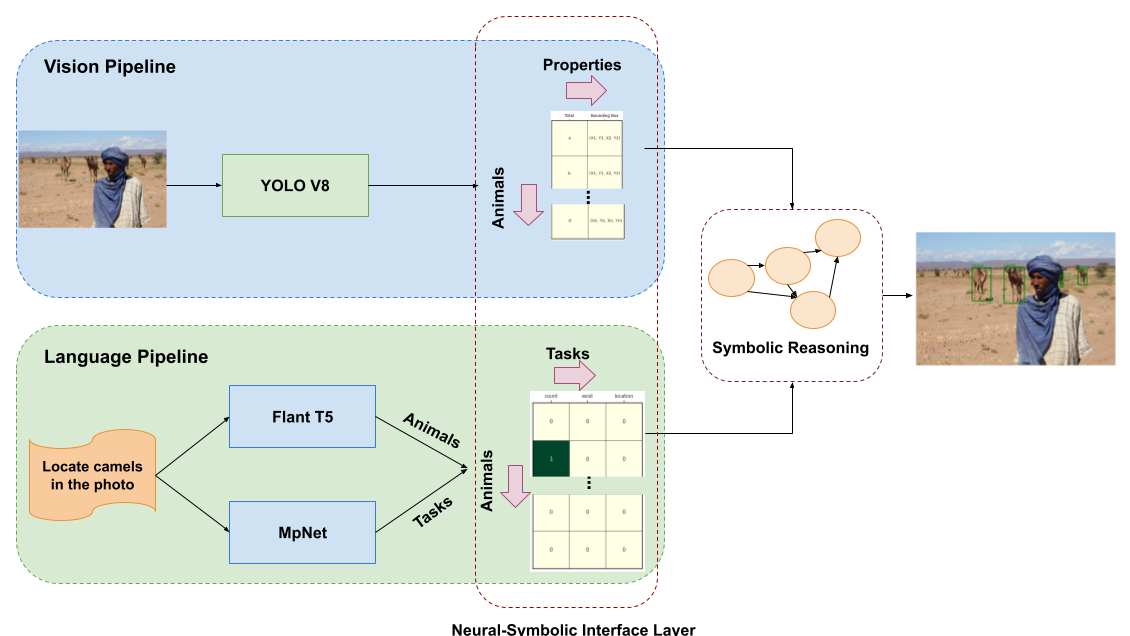}
  \end{center}
  \caption{Overview of ViLLa.}
  \label{fig:ViLLa}
\end{figure}

The ViLLa model integrates three interconnected modules: a vision pipeline, a language understanding pipeline, and a symbolic reasoning engine, shown in the figure \ref{fig:ViLLa}. The vision pipeline uses the YOLOv8 object detection model to identify animals within a given image and extract their class names and bounding boxes. This structured visual information is converted into symbolic way with animal names and its properties(count, bounding boxes), which are asserted into a Prolog-based knowledge base. The language pipeline uses a fine-tuned FLAN-T5 model to extract animal entities mentioned in natural language queries and the MPNet-based sentence transformer to classify the query task into predefined categories such as \textit{counting}, \textit{existence}, or \textit{location}. Once both visual and textual representations are symbolically grounded, the symbolic reasoning module interprets the query using logic rules and responds by retrieving or inferring information from the knowledge base. The pseudo code is shown in the algorithm \ref{alg:villa}

\begin{algorithm}
\caption{ViLLa: Vision-Language-Logic Approach}
\label{alg:villa}
\begin{algorithmic}
    \STATE \textbf{Input:} Image $I$, Natural Language Query $Q$
    \STATE \textbf{Output:} Logical Answer $A$

    \STATE $D \gets$ ObjectDetector.predict($I$)
    \FOR{each $d$ in $D$}
        \STATE Extract $animal$, $count$, and $bbox$ from $d$
        \STATE Assert facts to Prolog: $animal(a, c)$, $animal\_bbox(a, x1, y1, x2, y2)$
    \ENDFOR

    \STATE $A_{types} \gets$ AnimalModel.extract\_animals($Q$)
\STATE $I_{task} \gets$ TaskModel.detect\_task($Q$)

\IF{$I_{task}$ is count}
    \FOR{each $a$ in $A_{types}$}
        \STATE $A[a] \gets$ Query Prolog for $animal(a, c)$
    \ENDFOR
\ELSIF{$I_{task}$ is existence}
    \FOR{each $a$ in $A_{types}$}
        \STATE $A[a] \gets$ Query Prolog for $animal\_exists(a, c)$
    \ENDFOR
\ELSIF{$I_{task}$ is location}
    \FOR{each $a$ in $A_{types}$}
        \STATE $A[a] \gets$ Query Prolog for $animal\_bbox(a, x1, y1, x2, y2)$
    \ENDFOR
\ENDIF

\end{algorithmic}
\end{algorithm}

\subsection{Vision Pipeline}

To extract structured information from images, a YOLOv8-based object detection model was trained on a large-scale dataset. This model processes an input image and outputs both the object classes and their corresponding spatial locations in the form of bounding boxes.

Given an input image \( I \in \mathbb{R}^{H \times W \times 3} \), the model returns a set of detected objects:
\begin{equation}
\mathcal{D} = \left\{(c_i, \mathbf{b}_i) \mid i = 1, 2, \dots, N \right\}
\label{eq:detection_set}
\end{equation}
where
     \( c_i \in \mathcal{C} \): is the class label, \( \mathbf{b}_i \in \mathbb{R}^4 \): is the bounding box, \( N \): is the total number of detected objects in image \( I \)

Each detection is converted into a symbolic representation:
\begin{equation}
\texttt{symbol}_i = \left( c_i, \mathbf{b}_i \right)
\label{eq:symbol}
\end{equation}

To determine the number of animals per class, a class-specific counting function was defined as:
\begin{equation}
\texttt{count}(c) = \sum_{i=1}^{N} \mathbb{1}_{[c_i = c]}
\label{eq:count}
\end{equation}
where \( \mathbb{1}_{[c_i = c]} \): is an indicator function that evaluates to 1 if \( c_i = c \), and 0 otherwise
    \( \texttt{count}(c) \): gives the total number of detected objects belonging to class \( c \)

This vision pipeline thus provides a high-level symbolic interface between deep vision models and downstream reasoning components in neuro-symbolic system.

\subsection{Language Pipeline}

The language pipeline translates natural language queries into structured symbolic representations usable by a symbolic reasoning engine. It comprises two submodules: animal entity extraction and task classification.

\subsubsection{Animal Entity Extraction}
Given a query \( Q \),  Google's flan t5 large model \cite{chung2022scalinginstructionfinetunedlanguagemodels}  was used, a fine-tuned instruction-following variant of the T5 architecture, to extract animal class names. This model is prompted with a task-specific instruction to return the animal names mentioned in the query:

\begin{equation}
\mathcal{A}(Q) = \{ a_1, a_2, \dots, a_K \}
\label{eq:animal_list}
\end{equation}
where: \( a_k \in \mathcal{C} \): each animal class from a known class vocabulary \( \mathcal{C} \), \( K \): number of animals mentioned in the query

\subsubsection{Task Classification}

To classify the query task, The MPNet base v2 model was used from the SentenceTransformers library. MPNet encodes the query and a set of predefined labels (\texttt{counting}, \texttt{existence}, \texttt{location}) into fixed-length embeddings. Cosine similarity is computed between the embeddings:

\begin{equation}
\texttt{Task}(Q) = \arg\max_{l \in \mathcal{L}} \cos\left( \mathbf{e}_Q, \mathbf{e}_l \right)
\label{eq:task_class}
\end{equation}
where \( \mathcal{L} = \{\texttt{counting}, \texttt{existence}, \texttt{location} \} \): candidate task labels, \( \mathbf{e}_Q \) is query embedding generated by MPNet, \( \mathbf{e}_l \): embedding of label \( l \)

The final symbolic output of the language module is represented as:

\begin{equation}
\mathcal{S}_{\text{lang}} = (\mathcal{A}(Q), \texttt{TaskFlags}(Q))
\label{eq:language_output}
\end{equation}
where \( \texttt{TaskFlags}(Q) \) is a dictionary assigning a boolean value to each label in \( \mathcal{L} \), indicating whether the label is semantically active in the query.

This structured output provides a symbolic interface between natural language and logical reasoning.

\subsection{Symbolic Reasoning Module}
The final component of the framework is the symbolic reasoning module, which is responsible for interpreting and answering structured queries generated by the language and vision modules. This module is implemented using the Prolog logic programming language and is accessed through the Python PySwip interface.

Based on symbolic facts extracted from the image, such as animal classes, counts, and bounding boxes, and structured representations of textual queries, the module applies declarative logic rules to derive interpretable answers. Reasoning tasks span multiple query types, including count, existence, and spatial localization, and are extended here to handle multiple animals simultaneously, as well as multiple locations for the same animal class.

To retrieve the count of each animal class $a_i$ in a given query set $A = \{a_1, a_2, \dots, a_n\}$, the knowledge base is queried iteratively:

\[
\text{Count}(a_i) := \text{animal}(a_i, C_i) \quad \forall a_i \in A
\]

\noindent where $C_i \in \mathbb{N}$ denotes the number of instances detected for the animal $a_i$. This allows multiple animal types to be counted within a single query, returning a dictionary of class-count pairs.

To determine the presence of each animal $a_i \in A$ in the image, a Prolog rule evaluates the existence of each class:

\[
\text{Exists}(a_i) := \text{animal\_exists}(a_i, C_i) \quad \forall a_i \in A
\]

\noindent This returns a Boolean flag for each queried animal class, indicating whether it was found in the current image context. The logic supports multiple checks in parallel for efficient response to composite queries.

For spatial location queries, the bounding box coordinates of each instance of animal $a_i$ are retrieved from the knowledge base using the rule:

\[
\text{Location}(a_i) := \text{animal\_bbox}(a_i, X1, Y1, X2, Y2)
\]

\noindent Since an image can contain multiple instances of the same animal class, symbolic knowledge base supports multiple bounding-box facts for a single class. The query returns a list of bounding boxes for each class $a_i$, where each element is a tuple $(X1, Y1, X2, Y2)$. This allows the system to visually localize all detected instances of a given animal and overlay labeled boxes directly on the input image for interpretability.

\section{Results}

\begin{figure}[!t]
  \centering
  \includegraphics[width=0.8\linewidth]{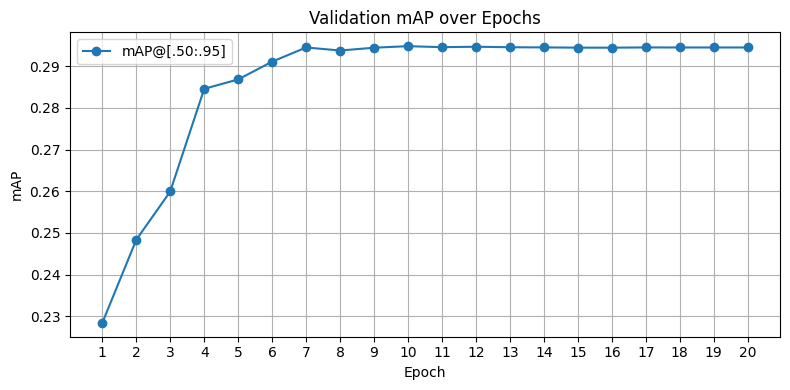}
  \caption{mAP over epochs on Yolo v8.}
  \label{fig:mAP}
\end{figure}

For this project, two publicly available animal image datasets from Kaggle were combined to ensure a broader and more representative coverage of animal types across various natural environments. The first dataset focused on wildlife in nature reserves across South Africa and contained labeled images of four key species-buffalo, elephant, rhino, and zebra—commonly found in those regions. To expand the diversity of animal classes and scenarios, it was merged with an object detection dataset sourced from Google’s Open Images V6+, which includes a variety of animals in different environments. This enhanced dataset added species such as  lion, leopard, cheetah, tiger, bear (including brown and polar bears), and others like butterfly, duck, camel, and crocodile, etc. The combined dataset allowed the system to operate on a wider set of animals and conditions for visual reasoning tasks.

\begin{figure}[!t]
  \centering
  \includegraphics[width=0.8\linewidth]{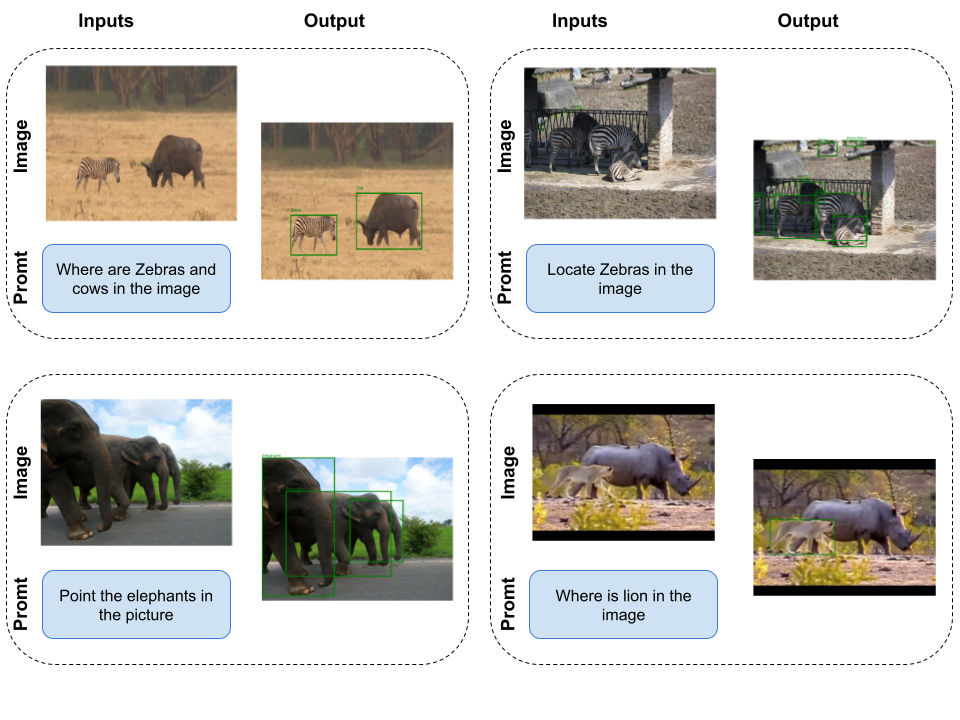}
  \caption{Qualitative outputs for the location task.}
  \label{fig:location_result}
\end{figure}

I trained a YOLOv8 model using the annotated animal dataset with a 70:10:20 train:validation:test split. The training was conducted on a system running Ubuntu 22.04 using NVIDIA Tesla T4 and K80 GPUs. The model was trained for 20 epochs with an mAP evalution. The performance of our model is shown in the figure \ref{fig:mAP}.

\begin{figure}[!t]
  \centering
  \includegraphics[width=0.8\linewidth]{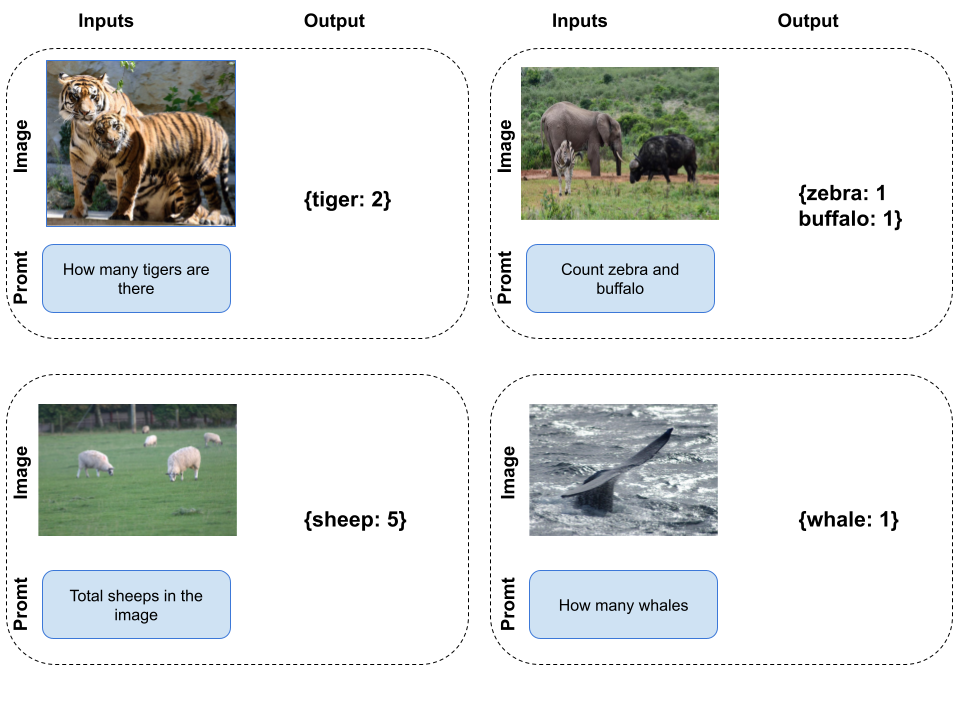}
  \caption{Qualitative outputs for the count task.}
  \label{fig:count_result}
\end{figure}

The results shown in Figures \ref{fig:location_result} and \ref{fig:count_result} highlight how the ViLLa framework supports different types of wildlife monitoring queries. The first set of examples (\ref{fig:location_result}) focuses on identifying where animals appear in an image. For questions like “Where are zebras and cows in the image?” or “Locate zebras in the image,” the system accurately finds the mentioned animals and places clear bounding boxes around them. It handles both single and multiple-object cases, even when the animals are partially hidden or grouped together, showing that the framework can manage real-world image complexity while remaining easy to interpret.

The second set (\ref{fig:count_result}) demonstrates how the system answers questions involving quantity, such as “How many tigers are there?” or “Count zebra and buffalo.” Without needing separate rules for each case, it reads the question, identifies the animals mentioned, and returns the correct count based on what it sees in the image. These examples include both common and uncommon animals, as well as mixed-species scenes, confirming that the system works reliably across different types of input.

Overall, these examples show how ViLLa brings together visual detection and simple rule-based logic to provide clear, structured answers. By combining object recognition with straightforward text understanding, it allows users to ask practical questions and get useful responses—without relying on complex, difficult-to-understand processing steps. The approach also makes it easy to extend the system later with more types of queries or information.
\section{Future Work}
`

In this work, A YOLO-based vision model was trained for animal object detection and paired it with pretrained language models to efficiently extract symbolic representations from both modalities. This allowed for a lightweight and interpretable system for animal monitoring tasks. However, the language components, although effective in zero-shot settings, were not fine-tuned for animal-specific text. In future studies, The plan is to further train these pretrained language models on domain-specific datasets to improve their semantic parsing and task classification, especially in challenging contexts such as complex query structures.

Currently, the system executes one task per query—such as counting or locating animals—based on identified task. A logical next step is to expand the framework to support multi-task learning. For example, a user query like “How many elephants are near the water?” could simultaneously trigger both counting and spatial reasoning modules. This would enable richer, more integrated responses and move the system closer to real-world applicability where complex, compound queries are common.

Furthermore, to advance the reasoning capabilities beyond basic logical rules, The aim is to include structured knowledge graphs \cite{zheng2024knowledge} \cite{tan2023literalawareknowledgegraphembedding}that represent relationships between animals, their habitats, and spatial co-occurrences. These graphs would act as a bridge between visual detections and textual interpretations, allowing the system to reason over contextual and relational knowledge, such as predator-prey dynamics or typical movement patterns. This upgrade to the symbolic reasoning backend would reduce reliance on black-box AI components and support more interpretable, semantically meaningful answers.

\section{Conclusion}

In this project, a novel neuro-symbolic framework, ViLLa, was proposed to integrate computer vision, natural language understanding, and symbolic reasoning for interpreting queries about animals in images. The system bridges the gap between visual perception, language input, and logical inference, enabling tasks such as object counting, presence verification, and spatial localization—all driven by natural language queries. Notably, ViLLa produces accurate and interpretable results without requiring additional training of the underlying models.

Overall, this work demonstrates the effectiveness of combining modular components with symbolic logic to develop flexible, transparent systems for structured visual reasoning. ViLLa offers a solid foundation for future extensions in multi-task querying, visual-language grounding, and interpretable decision-making—particularly in domains like wildlife monitoring, where clarity and adaptability are essential.

\bibliographystyle{unsrt}  
\bibliography{references}

\end{document}